\documentclass[conference]{IEEEtran}
\usepackage{graphicx}
\usepackage{bm}
\usepackage{amsmath}
\usepackage{amssymb}
\usepackage{esvect}
\usepackage{amsmath} %
\usepackage{amssymb}  %
\usepackage[mathscr]{euscript}
\usepackage{csvsimple}
\usepackage{pgfplotstable}
\usepackage{nicefrac}
\usepackage{tabularx}

  \newcommand{\T}{^\mathsf{T}}                  %
  \newcommand{\R}{\mathbb{R}}                   %
  \newcommand{\E}{\mathrm{E}}                   %
  \newcommand{\N}{\mathrm{N}}                   %
  \newcommand{\diff}{\,\mathrm{d}}              %
  \usepackage{bm}
  \newcommand{\mathbold}[1]{\bm{#1}}
  \newcommand{\mbf}[1]{\mathbf{#1}}
  \newcommand{\vect}[1]{\mbf{#1}}
  \newcommand{\vectb}[1]{\bm{#1}}

\newcommand{\vepsilon}[0]{\mathbold{\varepsilon}}

\newcommand{\vmu}[0]{\mathbold{\mu}}
\newcommand{\vomega}[0]{\mathbold{\omega}}
\newcommand{\vxi}[0]{\mathbold{\xi}}

\newcommand{\MSigma}[0]{\mathbold{\Sigma}}

\renewcommand{\mid}[0]{\,|\,}
\newcommand{\vzero}[0]{\mathbold{0}}
\newcommand{\veps}[0]{\mathbold{\varepsilon}}

\newcommand{\va}{\mbf{a}}
\newcommand{\vb}{\mbf{b}}

\newcommand{\ve}{\mbf{e}}
\newcommand{\vf}{\mbf{f}}
\newcommand{\vg}{\mbf{g}}
\newcommand{\vh}{\mbf{h}}

\newcommand{\vm}{\mbf{m}}

\newcommand{\vp}{\mbf{p}}
\newcommand{\vq}{\mbf{q}}
\newcommand{\vr}{\mbf{r}}

\newcommand{\vu}{\mbf{u}}
\renewcommand{\vv}{\mbf{v}}

\newcommand{\vx}{\mbf{x}}
\newcommand{\vy}{\mbf{y}}
\newcommand{\vz}{\mbf{z}}

\newcommand{\MC}{\mbf{C}}

\newcommand{\MF}{\mbf{F}}

\newcommand{\MH}{\mbf{H}}
\newcommand{\MI}{\mbf{I}}

\newcommand{\MK}{\mbf{K}}

\newcommand{\MP}{\mbf{P}}
\newcommand{\MQ}{\mbf{Q}}
\newcommand{\MR}{\mbf{R}}
\newcommand{\MS}{\mbf{S}}
\newcommand{\MT}{\mbf{T}}

  \newcommand{\eg}{\textit{e.g.}}
  
  \newcommand{\cf}{\textit{cf.}}
  \newcommand{\etc}{\textit{etc.}}
  \newcommand{\etal}{\textit{et~al.}}

  \usepackage{booktabs}

  \usepackage{tikz,pgfplots}
  \usetikzlibrary{plotmarks,shapes,arrows}
  \pgfplotsset{compat=newest} 
  \pgfkeys{/pgf/number format/.cd,1000 sep={}}

  \newlength\figureheight
  \newlength\figurewidth

  \usepackage[pagebackref=false,breaklinks=true,letterpaper=true,colorlinks,bookmarks=false, citecolor=black, linkcolor=black, urlcolor=black]{hyperref}

  \usetikzlibrary{external}
  \usepackage[capitalize,nameinlink]{cleveref}
  \crefname{section}{Sec.}{Secs.}
  \crefname{proposition}{Prop.}{Props.}
  \crefname{lemma}{Lem.}{Lems.}
  \crefname{model}{Mod.}{Mods.}
  \crefname{appendix}{App.}{Apps.}

  \makeatletter
  \let\NAT@parse\undefined
  \makeatother

  \makeatletter
  \let\NAT@parse\undefined
  \makeatother
  \usepackage[square,numbers,sort&compress]{natbib}

  \makeatletter
  \let\MYcaption\@makecaption
  \makeatother

  \usepackage[font=footnotesize]{subcaption}

  \makeatletter
  \let\@makecaption\MYcaption
  \makeatother

\definecolor{mycolor0}{rgb}{0.2667,0.4471,0.7098}
\definecolor{mycolor1}{rgb}{0.1647,0.6706,0.3804}
\definecolor{mycolor2}{rgb}{0.8275,0.2627,0.3059}
\definecolor{mycolor3}{rgb}{0.5216,0.4392,0.7176}
\definecolor{mycolor4}{rgb}{0.8118,0.7255,0.4118}
\definecolor{mycolor5}{rgb}{0.2745,0.7176,0.8157}

\definecolor{mylcolor0}{rgb}{0.6902,0.7686,0.8863}
\definecolor{mylcolor1}{rgb}{0.5451,0.8902,0.6941}
\definecolor{mylcolor2}{rgb}{0.9412,0.7490,0.7647}
\definecolor{mylcolor3}{rgb}{0.8627,0.8392,0.9176}
\definecolor{mylcolor4}{rgb}{0.9569,0.9373,0.8667}
\definecolor{mylcolor5}{rgb}{0.7529,0.9020,0.9373}
\definecolor{mylcolor6}{rgb}{0.8750,0.8750,0.8750}

  \definecolor{primarycolor}{HTML}{2C6EBA}
  \definecolor{secondarycolor}{HTML}{7CAC56}

\pgfplotsset{every axis/.append style={grid style={line width=0.6pt,dotted,gray}}}

\begin{document}

\title{A Look at Improving Robustness in Visual-inertial SLAM by Moment Matching}
\author{\IEEEauthorblockN{Arno Solin}
\IEEEauthorblockA{Aalto University / Spectacular AI\\
Espoo, Finland\\
\parbox{.3\textwidth}{\centering arno.solin@aalto.fi}}
\and
\IEEEauthorblockN{Rui Li}
\IEEEauthorblockA{Aalto University\\
Espoo, Finland\\
\parbox{.3\textwidth}{\centering rui.li@aalto.fi}}
\and
\IEEEauthorblockN{Andrea Pilzer}
\IEEEauthorblockA{Aalto University / FCAI \\
Espoo, Finland\\
\parbox{.3\textwidth}{\centering andrea.pilzer@aalto.fi}}
}

\maketitle

\begin{abstract}
  The fusion of camera sensor and inertial data is a leading method for ego-motion tracking in autonomous and smart devices. State estimation techniques that rely on non-linear filtering are a strong paradigm for solving the associated information fusion task. The \textit{de~facto} inference method in this space is the celebrated extended Kalman filter (EKF), which relies on first-order linearizations of both the dynamical and measurement model. This paper takes a critical look at the practical implications and limitations posed by the EKF, especially under faulty visual feature associations and the presence of strong confounding noise. As an alternative, we revisit the assumed density formulation of Bayesian filtering and employ a moment matching (unscented Kalman filtering) approach to both visual-inertial odometry and visual SLAM. Our results highlight important aspects in robustness both in dynamics propagation and visual measurement updates, and we show state-of-the-art results on EuRoC MAV drone data benchmark.
\end{abstract}

\section{Introduction}
\noindent
Visual inertial odometry (VIO) and simultaneous localization and mapping (SLAM) are leading methods for ego-motion tracking in autonomous and smart devices, such as drones, robots, and vehicles. Furthermore, the same technology is also the key enabler for inside-out tracking for augmented and virtual reality (AR/VR) in smartphones and headsets.

\begin{figure}[t!]
  \tiny
  \begin{tikzpicture}[outer sep=0]
    \node[minimum width=\columnwidth,minimum height=0.5625\columnwidth,rounded corners=2mm,path picture={
      \node at (path picture bounding box.center){
        \includegraphics[width=\columnwidth]{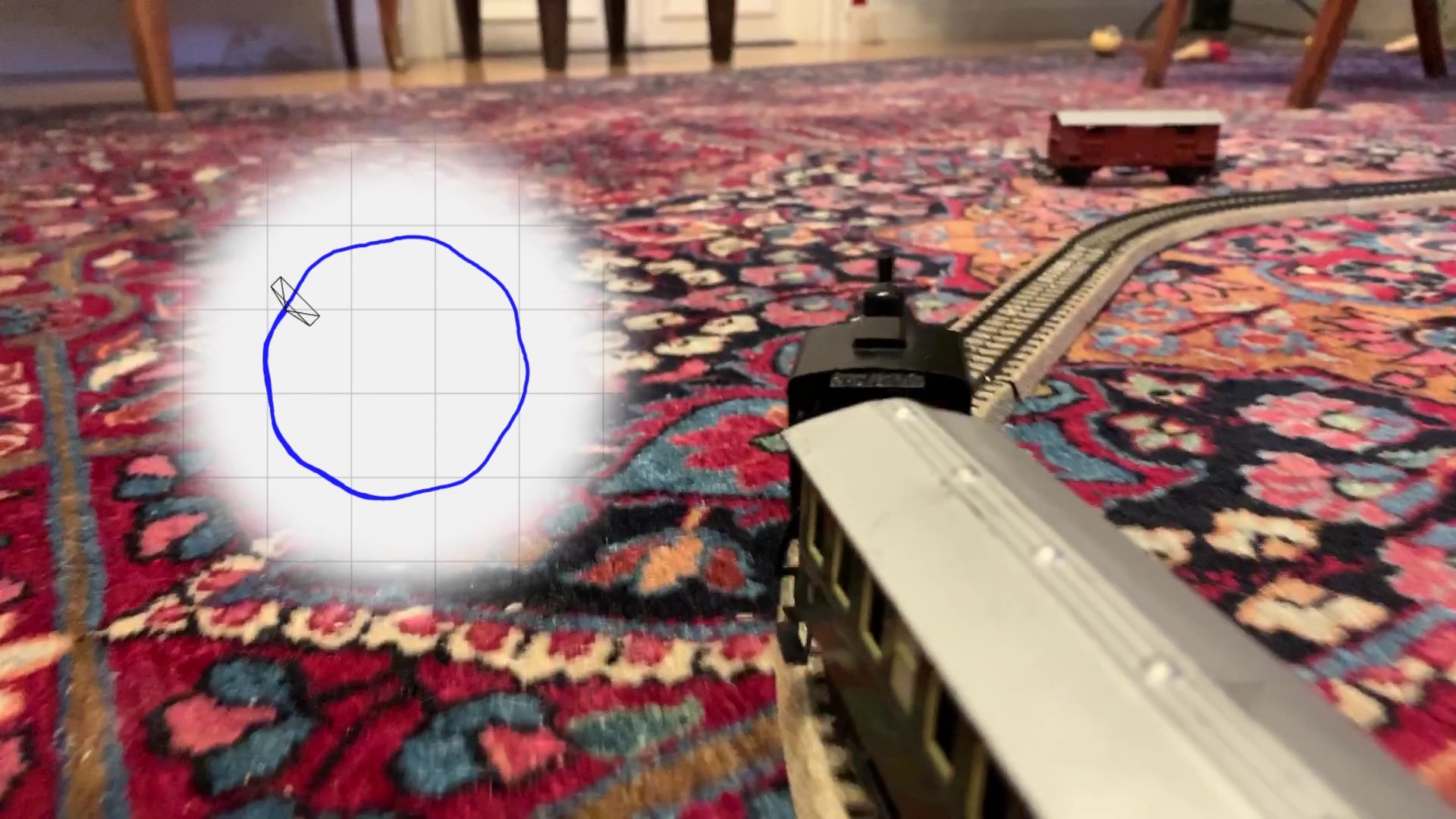}
      };}] at (0,0) {};
  \end{tikzpicture}
  \caption{Motivating example: A visual-inertial odometry setting with a smartphone camera attached to a model train car. The high-frequency jitter from the tracks and severe motion blur degrade both the IMU and camera data, which calls for principled inference methods. The overlaid track shows that the estimates do not drift away (grid 50~cm).}
  \label{fig:teaser}
\end{figure}

\cref{fig:teaser} shows an example ego-motion tracking task, where a smartphone with a camera and an IMU has been attached to a model train car. Both VIO and visual SLAM tackle the problem of motion estimation, but in VIO the methods typically do not have long-term memory, while in SLAM the central idea is position estimation and map building in tandem. Thus, VIO methods are typically employed in scenarios which are not constrained to or expected to stay in the same space, whereas SLAM methods are good for correcting drift in constrained spaces. 

Both these estimation problems can be formulated as a state estimation task, for which non-linear filtering (see, \eg, \cite{Sarkka:2013}) is a strong paradigm. However, due to computational (real-time) constraints and perhaps also historical burden, the method of choice for state estimation in widely available methods and implementations of VIO and visual SLAM are based on the celebrated extended Kalman filter (EKF, see, \eg, \cite{Bar-Shalom+Li+Kirubarajan:2001}), which relies on first-order linearizations of both the dynamical and measurement model. Especially in computer vision related state estimation literature, the EKF often seems to be the sole estimation method being considered.

This paper takes a critical look at the practical implications and limitations posed by the EKF, especially under faulty visual feature associations and presence of strong confounding noise. As an alternative, we revisit the assumed density formulation (ADF, \cite{Maybeck:1982b,Ito+Xiong:2000,Wu+Hu+Wu+Hu:2006}) of Bayesian filtering and employ a moment matching (unscented Kalman filtering, \cite{Julier+Uhlmann+Durrant-Whyte:1995,Julier+Uhlmann+Durrant-Whyte:2000}) approach to both visual-inertial odometry and visual SLAM. We also discuss wider links to approximate inference methods in statistical machine learning.

The contributions of this paper are the following. \emph{(i)}~We revisit the assumed density filtering literature and provide a unifying view that can be leveraged in VIO and visual SLAM; \emph{(ii)}~we study aspects of robustness in an EKF-SLAM/ADF-SLAM updates; and \emph{(iii)}~we show state-of-the-art results with ADF predictions on the {\em EuRoC MAV drone flying benchmark} data set for online stereo tracking.

\section{Related Work}
\label{sec:related}
\noindent
EKF-SLAM~\cite{Bailey+Nieto+Guivant+Stevens+Nebot:2006,Barrau+Bonnabel:2015,MSCKF,paz2008divide} is a direct approach for formulating SLAM in a state estimation context. The idea in EKF-SLAM is to include the map in the system state and solve the localization and mapping problem jointly. This leads to a high-dimensional state, which typically renders sequential Monte Carlo methods (particle filtering) inefficient and encourages the use of alternative estimation methods. The problems related to local linearization in EKF are widely known, and the UKF has been employed in an UKF-SLAM fashion in the past (see, \eg, \cite{wang2007upf,huang2009complexity}). These methods have, however, never become mainstream, mostly due to other methods such as graph optimization based SLAM largely replacing state estimation methods in this context.

In visual-inertial odometry, state estimation methods have provided a successful branch of methods. These methods can be characterized to stem from the MSCKF~\cite{MSCKF} line of development which includes, \eg, PIVO~\cite{PIVO} and HybVIO~\cite{HybVIO}. HybVIO holds the current state-of-the-art results on public benchmark data sets for VIO in online mode. Other recent methods belonging to the same class include the hybrid-EKF-SLAM~\cite{hybrid-ekf-slam} method LARVIO~\cite{LARVIO} and S-MSCKF~\cite{stereoMSCKF}, which extends the original MSCKF to stereo cameras. These methods, following their EKF-SLAM predecessors~\cite{monoSLAM}, use an EKF to keep track of the VIO state.

From an inference methodology point of view, it is interesting how little attention alternative non-linear filtering methods have gained in the VIO context. Especially in the case of challenging benchmarks, sensitivity to outliers, general robustness, and well-calibrated uncertainties should be important. In machine learning, a wide variety of approximative inference methods have been considered, with the current gold-standard being various sampling schemes (see \cite{Gelman+Carlin+Stern+Dunson+Vehtari+Rubin:2013} for an overview), variational methods \cite{Wainwright-Jordan:2008,Opper+Archambeau:2009,Titsias:2009}, and expectation propagation (EP, \cite{minka2001expectation,jylanki2011robust, Kuss2006,Bui:2017}). All of these can also be formulated into a state estimation setting (see, \eg, discussion \cite{wilkinson2020,wilkinson2021bayes}). Inspired by the original formulation of EP~\cite{minka2001expectation} with foundations in filtering and smoothing, we draw from the motivation in \cite{wilkinson2020}, and explore focus on moment matching in an assumed density context for improving inference in visual SLAM and VIO.

\section{Methods}
\label{sec:methods}
\noindent
In the following, we give an overview of an assumed density filtering (ADF) view into state estimation that we will use in the experiments. This framework has extensively been leveraged in signal processing and sensor fusion in the past, and it is summarised here mostly to give a consistent outlook of the employed methodology in the experiments that falls under the formulation of the model in \cref{sec:models} and the ADF inference framework.

\subsection{Models}
\label{sec:models}
\noindent
The model describes the evolution of the system dynamics over time steps $t_k \in \mathbb{R}$ for discrete steps $k=1,2,\ldots,T$, and couples the  latent (unobservable) dynamics to measurements through an observation (likelihood) model:
\begin{align}
  \vx_k &= \vf_k(\vx_{k-1},\veps_k), && \text{(prior)} \label{eq:dynamics}\\
  \vy_k &= \vh_k(\vx_k) + \vr_k, && \text{(likelihood)} \label{eq:measurement}
\end{align}
where $\vx_k \in \R^d$ is a vector-valued state variable, $\vf_k(\cdot,\cdot): \R^{d\times s} \to \R^d$ is a dynamical non-linear mapping from the previous state to the next state over time (note the time-dependency), and $\veps_k \sim \mathrm{N}(\vzero,\MQ_k)$ is an $s$-dimensional Gaussian process noise. 

For convenience, we consider additive noise models in \cref{eq:measurement}. The likelihood, $p(\vy_k) = \mathrm{N}(\vy_k \mid \vh_k(\vy_k), \MR_k)$, is given by a non-linear mapping $\vh_k(\cdot): \R^d \to \R^m$ that maps the latent states to observations, under corrupting Gaussian noise $\vr_k \sim \mathrm{N}(\vzero,\MR_k)$.

In practice, performing probabilistic inference on the latent state in models of the above can be challenging. In case of both $\vf(\cdot)$ and $\vh(\cdot)$ being linear mappings, the problem reduces to a linear-Gaussian state estimation problem, for which the Kalman filter is an optimal estimator. Under more general noise or, as in our case, non-linearities, no closed-form solution exists and one needs to resort to approximate inference methods. In practice, one can either employ sequential Monte Carlo (SMC, particle filtering) methods that characterize the state $\vx_k$ through a finite set of sample points, or one can use assumed density filtering methods, where the state is restricted to have a simple form (typically Gaussian). In the following sections, we explore the latter option.

\subsection{General Gaussian filtering}
\label{sec:gaussian-filter}
\noindent
We view the information fusion task under the perspective of general Gaussian filtering, where we consider the state $\vx_k$ to be Gaussian
  $\vx_k \triangleq \mathrm{N}(\vx_k \mid \vm_k, \MP_k)$.
To unify views from various Gaussian approximations to non-linear transforms---that give rise to various non-linear filtering schemes in information fusion---one can view them as different ways of approximating the Gaussian integrals of the form (see, \eg, presentation in \cite{Sarkka:2013})
\begin{equation}\label{eq:gauss-int}
  \E_{\mathrm{N}}[\bullet] 
  = \int [\bullet] \, \N(\vx \mid \vm,\MP) \diff \vx.
\end{equation}
For the dynamical model in \cref{eq:dynamics}, this gives a Gaussian assumed density filter (ADF) formulation (see \cite{Maybeck:1982b,Ito+Xiong:2000,Wu+Hu+Wu+Hu:2006}), where the prediction (time update) steps of the non-additive noise Gaussian (Kalman) filter are given as follows.
\begin{align}
  \vm_k^- &= \int \int \vf_k(\vx_{k-1},\veps_k) \, \nonumber \\ &\quad 
    \mathrm{N}(\vx_{k-1} \mid \vm_{k-1}, \MP_{k-1}) \, \mathrm{N}(\veps_k \mid \vzero,\MQ_k) \diff\veps_k \diff\vx_{k-1}, \label{eq:pred-mean-int} \\
  \MP_k^- &= \int \int (\vf_k(\vx_{k-1},\veps_k) - \vm_k^-)\,(\vf_k(\vx_{k-1},\veps_k) - \vm_k^-)\T \nonumber \\ &\quad
  \mathrm{N}(\vx_{k-1} \mid \vm_{k-1}, \MP_{k-1}) \, \mathrm{N}(\veps_k \mid \vzero,\MQ_k) \diff\veps_k \diff\vx_{k-1}, \label{eq:pred-cov-int}
\end{align}
where $\vx_k^- = \mathrm{N}(\vx_k^- \mid \vm_k^-, \MP_k^-)$ represents the predictive distribution. Similarly, we write down the corresponding additive noise ADF update equations as follows.
\begin{align}
  \vmu_k &= \int \vh_k(\vx_k) \, \mathrm{N}(\vx_k \mid \vm_k^-, \MP_k^-) \diff\vx_k \label{eq:inn-mean-int} \\
  \MS_k &= \int (\vh_k(\vx_k) - \vmu_k) \, (\vh_k(\vx_k) - \vmu_k)\T  \nonumber \\ &\qquad\qquad\qquad \mathrm{N}(\vx_k \mid \vm_k^-, \MP_k^-) \diff\vx_k + \MR_k, \label{eq:inn-cov-int} \\
  \MC_k &= \int (\vx_k - \vm_k^-) \, (\vh_k(\vx_k) - \vmu_k)\T  \nonumber \\ &\qquad\qquad\qquad \mathrm{N}(\vx_k \mid \vm_k^-, \MP_k^-) \diff\vx_k, \label{eq:cross-int}
\end{align}
where $\vmu_k$ and $\MS_k$ are called the innovation mean and covariance respectively, and $\MS_k$ is a cross-covariance term. The Kalman {\em update step} then becomes:
\begin{equation}
\begin{split} \label{eq:update}
  \MK_k &= \MC_k \, \MS_k^{-1}, \\
  \vm_k &= \vm_k^- + \MK_k \, (\vy - \vmu_k),  \\
  \MP_k &= \MP_k^- - \MK_k \, \MS_k^{-1} \, \MK_k\T,
\end{split}
\end{equation}
where $\vy_k$ is an observation vector of measurements, $\MK_k$ is a gain term, and the filtering solution at time step $t_k$ is given by $\vx_k = \mathrm{N}(\vx_k \mid \vm_k, \MP_k)$. 
 
Now we see that the integrals in \cref{eq:pred-mean-int,eq:pred-cov-int,eq:inn-mean-int,eq:inn-cov-int,eq:cross-int} are all of the Gaussian integral (expectation) form of \cref{eq:gauss-int}. Thus this framework gives rise to various non-linear filtering schemes depending on how the (typically intractable) integrals are solved. Various quadrature (or cubature \cite{Cools:1997}) methods give rise to, \eg, the Gauss--Hermite Kalman filter (GHKF), the cubature Kalman filter (CKF), the unscented Kalman filter (UKF), and so on.

The ADF framework effectively performs moment matching (for finding the first two moments, mean and covariance, that characterize the Gaussian). The general ADF framework in its basic form thus links directly to single-sweep expectation propagation (EP) methods in machine learning.

\subsection{Inference by extended Kalman filtering}
\label{sec:EKF}
\noindent
The general Gaussian filtering methodology in \cref{sec:gaussian-filter} can also give rise to an interpretation of the extended Kalman filter, where the integrals of form \cref{eq:gauss-int} are solved through low-order Taylor approximations (first-order linerization). The EKF {\em prediction step} becomes:
\begin{equation} \label{eq:prediction_step}
\begin{split}
  \vm_k^- &= \vf_k(\vm_{k-1}, \vzero), \\
  \MP_k^- &= \MF_\vx \, \MP_{k-1} \, \MF_\vx\T + 
    \MF_{\veps} \, \vect{Q}_k \, \vect{F}_{\veps}\T,
\end{split}
\end{equation}
where the dynamic model is evaluated with the outcome from the previous step and zero noise, and $\MF_\vx(\cdot)$ denotes the Jacobian matrix of $\vf_k(\cdot, \cdot)$ with respect to $\vx$ and $\MF_{\veps}(\cdot)$ with respect to the process noise $\veps$.

The innovation terms and cross-covariance for the EKF {\em update step} can be approximated as follows
\begin{equation}
\begin{split}
  \vmu_k &= \MH_\vx \, \vm_k^- \\
  \MS_k  &= \MH_\vx \, \MP_k^- \, \MH_\vx\T + \MR_k,  \\
  \MC_k  &= \MP_k^- \, \MH_\vx\T,
\end{split}
\end{equation}
where $\MH_\vx$ denotes the Jacobian of the measurement model $\vh_k(\cdot)$ with respect to the state variables $\vx$. The rest of the update follows the general update equations in \cref{eq:update}.

\begin{figure}[t!]
  \footnotesize\centering
  \setlength{\figurewidth}{\columnwidth}
  \setlength{\figureheight}{\figurewidth}
  \def\datapath{./fig/}
  \input{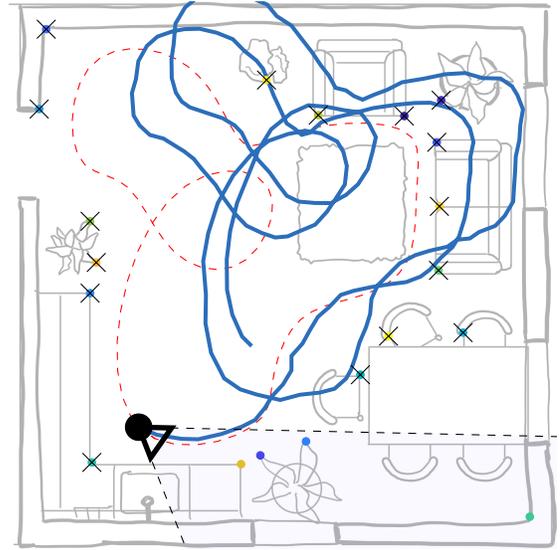}
  \caption{Simulation setup for the SLAM experiment, where a camera frustrum (black) observes noisy 1D projections of 2D feature points (coloured points visualized at their ground-truth locations) while moving along the dashed ground-truth path. The odometry is noisy and drifts, which would result in the solid blue path.}
  \label{fig:odometry}
  \vspace*{-1em}
\end{figure}

\subsection{Moment matching gives rise to the unscented Kalman filter}
\label{sec:UKF}
\noindent
Even if local linearization is efficient and often produces good results, if the non-linearities are not severe and the linearization point is reasonable, we argue that being able to match the moments more accurately can be beneficial. 

An alternative way of constructing an assumed density approximation to $p(\vx_{1:T})$ is to directly match the moments by solving the Gaussian integrals in \cref{eq:pred-mean-int,eq:pred-cov-int,eq:inn-mean-int,eq:inn-cov-int,eq:cross-int} by Gaussian quadrature methods. The approximation to \cref{eq:gauss-int} would take the form 
\begin{align}
  \E_{\mathrm{N}(\vmu, \MSigma)}[\vg(\vx)] 
  &= \int \vg(\vx) \, \N(\vx \mid \vmu,\MSigma) \diff \vx \nonumber \\
  &\approx \sum_i w_i \, \vg(\vz_i),
\end{align}
for an arbitrary integrand $\vg(\vx)$, weights $w^{(i)}$, and so called sigma points $\vz_i = \vmu + \sqrt{\MSigma} \, \vxi_i$. Here $\sqrt{\MSigma}$ denotes a square-root factor of $\MSigma$ such as the Cholesky decomposition. The multi-dimensional Gaussian quadrature (or {\em cubature}, see \cite{Cools:1997}) rule is characterized by the evaluation points and their associated weights $\{(\vxi_i,w_i)\}$.

\begin{figure*}[t!]
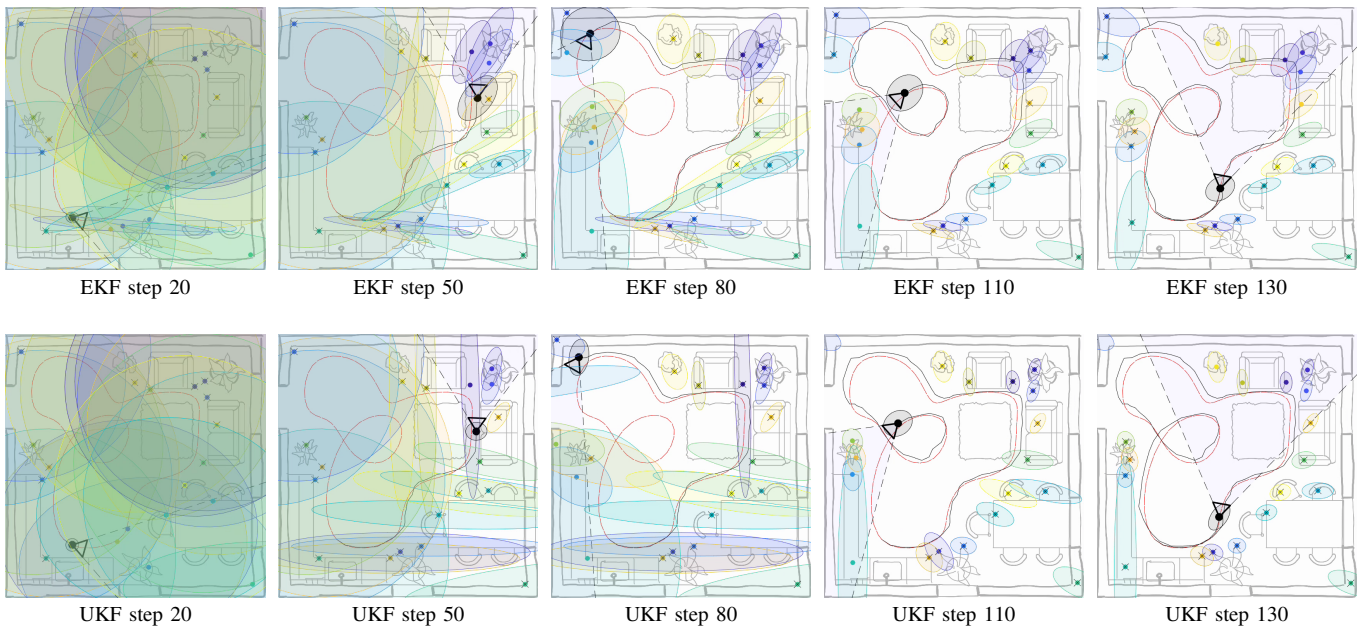

  \footnotesize
  \setlength{\figurewidth}{.2\textwidth}
  \setlength{\figureheight}{\figurewidth}
  \begin{tikzpicture}[outer sep=0]  
    \foreach \x/\t [count=\i] in {02/20,05/50,08/80,11/110,13/130} {
      \node[align=center] at (\i\figurewidth,0) {\includegraphics[width=.95\figurewidth,trim=250 95 200 65,clip]{fig/frames/ekf-\x}\\ EKF step \t};
      
      \node[align=center] at (\i\figurewidth,-1.2\figureheight) {\includegraphics[width=.95\figurewidth,trim=250 95 200 65,clip]{fig/frames/ukf-\x}\\ UKF step \t};
    }
  \end{tikzpicture}
  \caption{Sketch of linearization-based (EKF) and moment matching based (UKF) SLAM in a simulated two-dimensional setting. The marginal distributions extracted from the state for both the camera position and the feature positions are visualized by uncertainty ellipses (95\% credible intervals). Under no outliers and mild noise, both methods perform comparably well---which sets the starting point for subsequent experiments.}
  \label{fig:slam}
\end{figure*}

In machine learning, a typical choice for a numerical method for matching the moments would be Gauss--Hermite quadrature, which factorizes over the input dimensions leading to an exponential number ($p^d$) of function evaluations/sigma points in the input dimensionality $d$ for a desired order $p$. However, as the latent is typically high-dimensional in this type of sensor fusion tasks, and in order to guarantee scalability, we employ a symmetric 3\textsuperscript{rd} order cubature rule \cite{Arasaratnam+Haykin:2009} which similarly to Gauss--Hermite ($p=3$) is exact for polynomials up to degree 3. The points are given by scaled unit coordinate vectors $\ve_i$ such that
\begin{equation}
  \vxi_i = 
  \begin{cases} 
    \phantom{-}\sqrt{d}\,\ve_i, \quad \text{for~}i=1,\ldots,d, \\  
    -\sqrt{d}\,\ve_i, \quad \text{for~}i=d+1,\ldots,2d,
  \end{cases}
\end{equation}
and the associated weights are $w_i = \frac{1}{2d}$. This scheme gives rise to the so-called {\em cubature Kalman filter} (CKF, \cite{Arasaratnam+Haykin:2009}), which is a special case (under specific choice of transformation parameters) of the well-known {\em unscented Kalman filter} (UKF, \cite{Julier+Uhlmann+Durrant-Whyte:1995,Julier+Uhlmann+Durrant-Whyte:2000}). \looseness-1

In the non-additive prediction step, the double integral can be handled by augmenting the state with the noise process: $\hat{\vx} = (\vx, \veps) \in \R^{d+s}$. This means in practice that the augmented mean becomes $\hat{\vm} = (\vm_k, \vzero)$, the augmented covariance $\hat{\MP} = \mathrm{blkdiag}(\MP_k, \MQ_k)$---thus requiring $2(d+s)$ quadrature points.

In what follows, we refer to this approximate inference methodology as `UKF'. Depending on viewpoint it could also be referred to as the CKF, moment matching, or single-sweep expectation propagation (EP).

\section{Experiments}
\label{sec:experiments}
\noindent
We consider two different models in our experiments. In \cref{sec:EKF-SLAM}, we empirically study practical aspects of robustness under an EKF vs.\ UKF estimation setting in a simulated SLAM setup (for full control of the experiment). This experiment considers linear state dynamic and a non-linear measurement model. The second experiment setup in \cref{sec:VIO}, considers replacing the time update in HybVIO~\cite{HybVIO}, a state-of-the-art VIO model, with a UKF prediction step for improved robustness in real-world visual-inertial state estimation.

\subsection{Simulated EKF-SLAM vs.\ UKF-SLAM}
\label{sec:EKF-SLAM}
We revisit the classical EKF-SLAM idea, where both the camera pose (translation and orientation) and the relative world coordinate positions of visual map points are in the state of the system. This setting is sketched out in \cref{fig:odometry}, where the camera is represented with a camera cone (frustrum) that also shows the field of view. The ground truth positions of visual feature points are marked with coloured dots (the ones not currently visible in the field-of-view are overlaid with crosses). The system state is thus given by
\begin{equation}
  \vx_k = (\vectb{\pi}_k, \vp_k^{(1)},\vp_k^{(2)},\ldots,\vp_k^{(p)}),
\end{equation}
where $\vectb{\pi}_k$ is the camera pose (translation and rotation), and the $\vp_k^{(i)}$ is the current position of the feature point $\#i$. The SLAM learning task is thus balancing between exploiting the already learned feature point locations for estimating the camera location vs.\ learning the feature point locations with the help of the relative movement of the camera.

\begin{figure*}[t!]
  \scriptsize
  \setlength{\figurewidth}{.35\textwidth}
  \setlength{\figureheight}{.385\textwidth}
  \pgfplotsset{xtick={0,0.05,0.1,0.15},xticklabels={0\%,5\%,10\%,15\%}}
  \begin{subfigure}[b]{.15\textwidth}
    \centering
    \includegraphics[width=.95\textwidth,trim=250 95 200 65,clip]{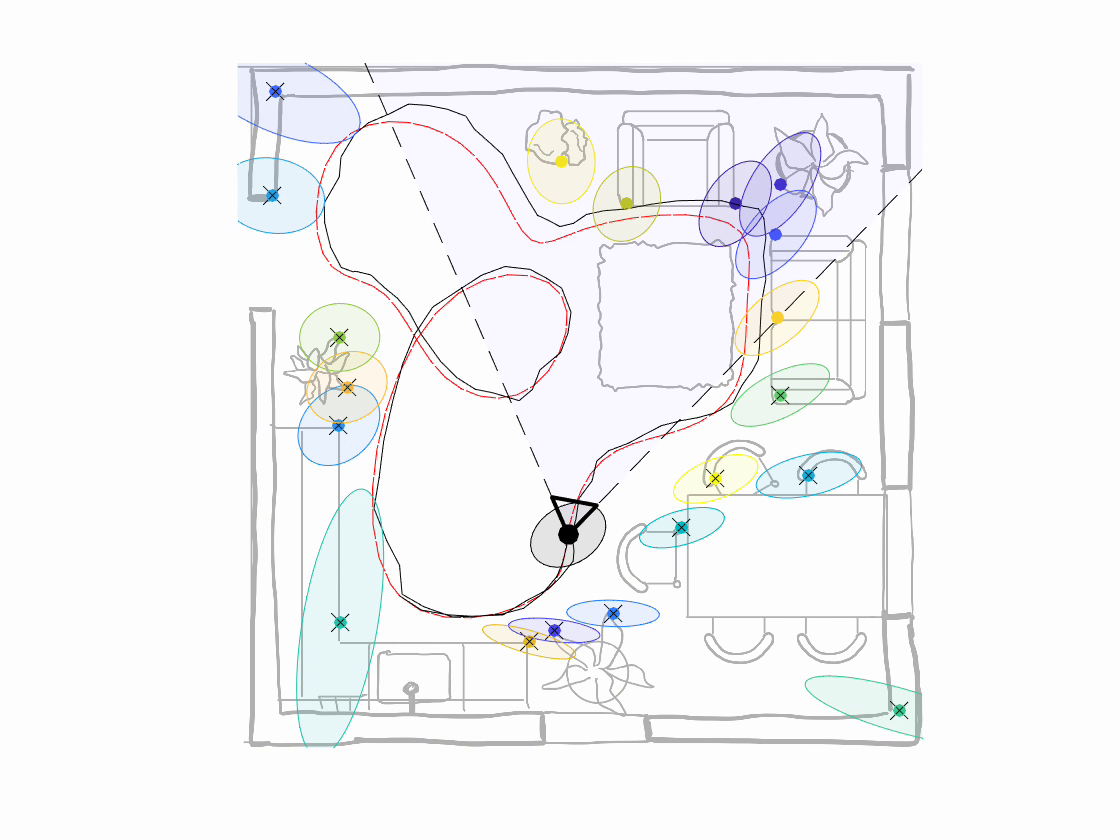} \\
    EKF step 130 \\[1em]
    \includegraphics[width=.95\textwidth,trim=250 95 200 65,clip]{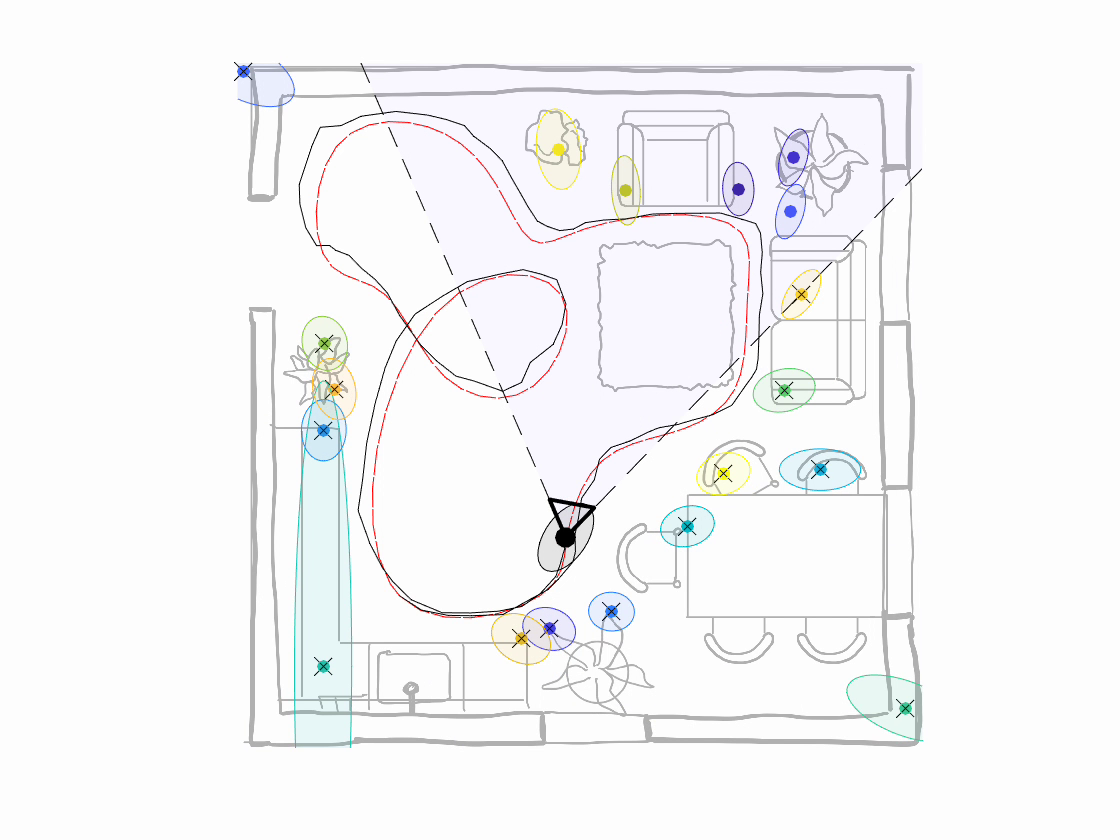} \\
    UKF step 130
    \captionsetup{justification=centering}
    \caption{No wrong \\ assignments}
  \end{subfigure}%
  \hfill
  \begin{subfigure}[b]{.32\textwidth}
    \centering
    \input{./fig/shuffle-1.tex}
    \caption{Normalized RMSE in path \\ ~}
  \end{subfigure}
  \hfill
  \begin{subfigure}[b]{.32\textwidth}
    \centering
    \input{./fig/shuffle-2.tex}
    \caption{Normalized RMSE of map after alignment \\ ~}    
  \end{subfigure}  
  \hfill
  \begin{subfigure}[b]{.15\textwidth}
    \centering
    \includegraphics[width=.95\textwidth,trim=250 95 200 65,clip]{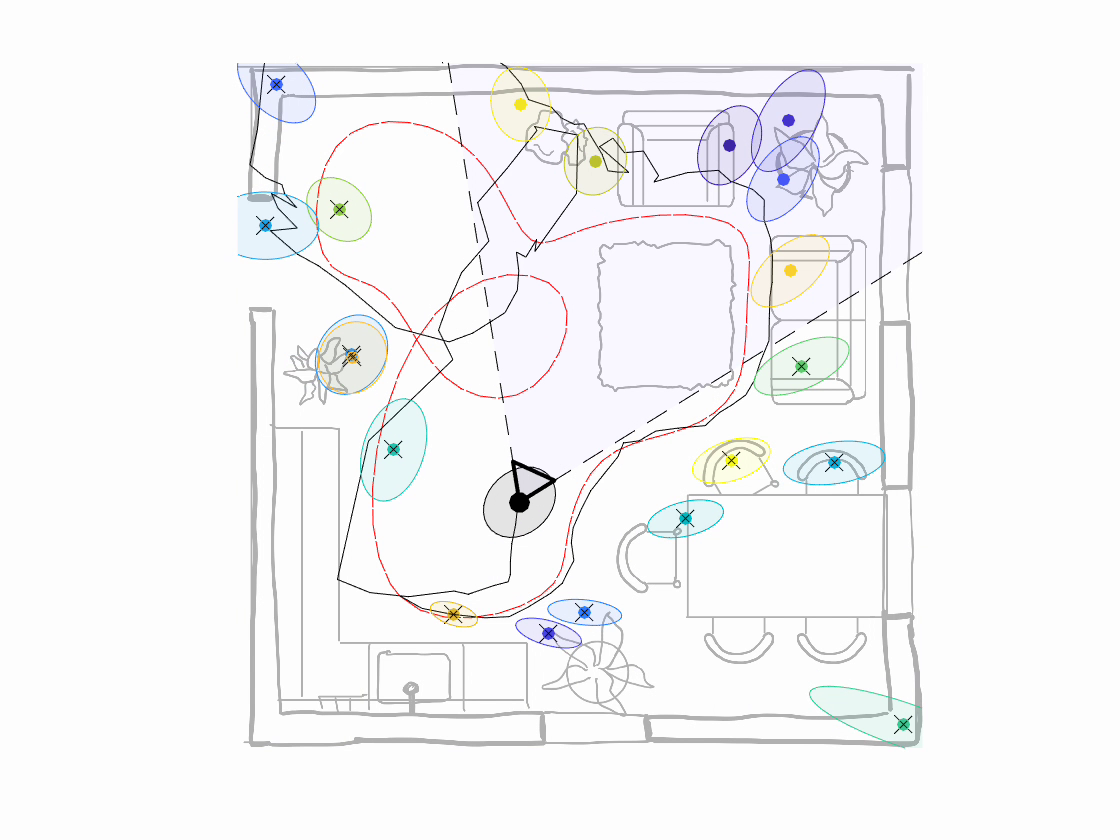} \\
    EKF step 130 \\[1em]
    \includegraphics[width=.95\textwidth,trim=250 95 200 65,clip]{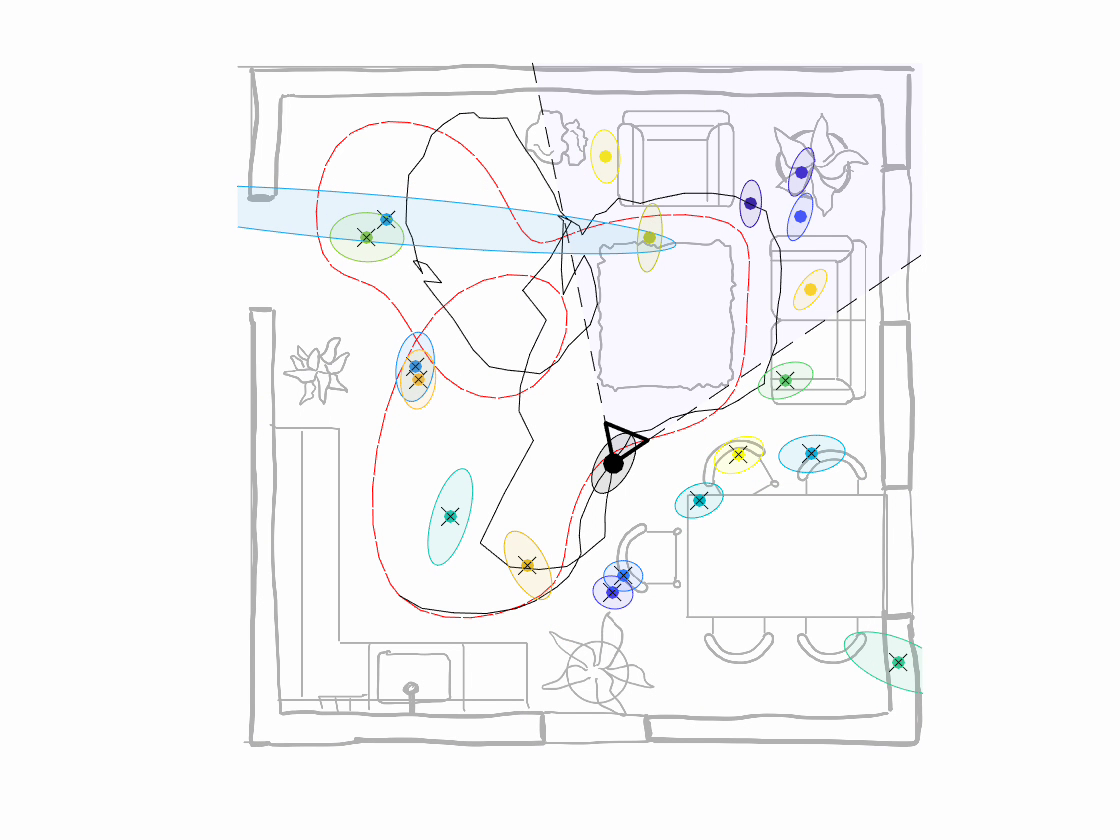} \\
    UKF step 130
    \captionsetup{justification=centering}    
    \caption{15\% wrong \\ assignments}    
  \end{subfigure}%
  \vspace*{-6pt}
  \caption{Simulated SLAM under erroneous feature point assignments. Panel (a) shows solutions for linearization (EKF) and moment matching (UKF) based SLAM under no wrong assignments. Panels (b)--(c) show the effect on RMSE over the path (b) and map feature point locations (c) after a procrustes alignment to align the scale based on the learned map. In (b), we visualize the RMSE as a function of proportion of corrupted feature point assignments (nearby features swapped). Each point corresponds to a full run of the SLAM and we visualize the moving average (solid line) and moving standard deviation (coloured patch). Moment matching appears  more stable than linearization. Panel (d) shows example results under $\sim$15\% wrong feature assignments.}
  \label{fig:wrong-slam}
  \vspace*{-6pt}
\end{figure*}

In geometric computer vision (\eg, \cite{Hartley+Zisserman:2003}), a camera projection model is characterized by so called {\em extrinsic} (external) and {\em intrinsic} (internal) camera parameters. The extrinsic parameters denote the coordinate system transformations from world coordinates to camera coordinates, while the intrinsic parameters map the camera coordinates to image coordinates. In a {\em pinhole camera} model, a 3D world this corresponds to
\begin{equation}\label{eq:camera}
  \begin{pmatrix} u & v & 1 \end{pmatrix}\T \propto \MK^\mathrm{cam} \begin{pmatrix} \MR\T & -\MR\T \vp \end{pmatrix} \begin{pmatrix} x & y & z & 1 \end{pmatrix}\T,
\end{equation}
where $(u,v)$ are the image (pixel) coordinates, $\vp_\mathrm{xyz} = (x,y,z) \in \R^3$ are the world coordinates, $\MK^\mathrm{cam}$ is the intrinsic matrix (with camera focal lengths and principal point), and the $\vp \in \R^3$ and  $\MR$ describe the position of the camera centre and the orientation in world coordinates respectively. From \cref{eq:camera}, given a set of fixed world coordinates and a known motion between frames, changes in pixel values $(u,v)$ are driven by the camera pose $\vectb{\pi}=\{\vp,\MR\}$.

For better control of the experiment and visualization, we simplify this model to a two-dimensional setting (where `images' are 1D projections of the 2D world points). This gives us the following measurement model ($y_k = h_k(\vx_k)+r_k$ as in \cref{eq:measurement})
\begin{align}
  u_k^{(i)} &= 
  \begin{pmatrix} f & c \\ 0 & 1 \end{pmatrix} \,
  \begin{pmatrix} \MR(\theta_k)\T & -\MR(\theta_k)\T\vp_k \end{pmatrix} \,
  \begin{pmatrix} \vp_k^{(i)} \\ 1 \end{pmatrix}, \nonumber \\
  y_k^{(i)} &= u_k^{(i)}(1) / u_k^{(i)}(2) + r_k, \label{eq:slam-meas}
\end{align}
where $y_k^{(i)}$ is the observed projection of feature point $\#i$ in the 1D `image', $f$ is a focal length parameter (we use $f=\nicefrac{3}{2}$), $c$ is the principal point parameter (we use $c=0$), $\MR(\theta_k)$ is a $2{\times}2$ rotation matrix from camera to world coordinates evaluated with state orientation variable $\theta_k$, $\vp_k$ is the translation state of the camera, and $\vp_k^{(i)}$ is the translation state of feature point~$\#i$. 

In this experiment, we will focus on the visual update part (as in \cref{eq:slam-meas}), and thus we consider a linear-Gaussian dynamic model (simplifying \cref{eq:dynamics}) with odometry increments for the pose ($\vectb{\pi}_k = \{\vect{p}_k, \theta_k\}$):
\begin{equation}\label{eq:slam-dyn}
  \begin{pmatrix} \vp_k \\ \theta_k \end{pmatrix} = 
  \begin{pmatrix} \vp_{k-1} \\ \theta_{k-1} \end{pmatrix} +
  \begin{pmatrix} \Delta\vp_{k} \\ \Delta\theta_{k} \end{pmatrix} +
  \vectb{\varepsilon}_k,
\end{equation}
where $\vu_k = (\Delta\vp_{k}, \Delta\theta_{k})$ can be considered a (noisy) control signal and $\vectb{\varepsilon}_k$ is a Gaussian noise process. The feature point locations $\vp_k^{(i)}$ obey unit dynamics. Just directly integrating \cref{eq:slam-dyn} would lead to a reconstruction of an odometry path as visualized in \cref{fig:odometry}---showing both noisy increments and severe drift.

As the dynamics are linear-Gaussian, we use a linear Kalman prediction step. For the update, we implement both an EKF update (linearization as in \cref{sec:EKF}) and compare that to moment matching as described in \cref{sec:UKF}. \cref{fig:slam} demonstrates linearization-based (EKF) and moment matching based (UKF) SLAM in our simulated two-dimensional setting. The marginal distributions extracted from the state for both the camera translation $\vp_k$ and the feature positions $\{\vp_k^{(i)}\}_{i=1}^{20}$ are visualized by uncertainty ellipses (95\% credible intervals). In \cref{fig:slam}, under mild noise and no outliers both approaches perform similarly. The experiment setting considers two loops around the space (to highlight map re-use), with 197 time steps in total. \looseness-1

\begin{figure*}[t!]
  \scriptsize
  \setlength{\figurewidth}{.35\textwidth}
  \setlength{\figureheight}{.385\textwidth}
  \begin{subfigure}[b]{.15\textwidth}
    \centering
    \includegraphics[width=.95\textwidth,trim=250 95 200 65,clip]{fig/frames/ekf-13} \\
    EKF step 130 \\[1em]
    \includegraphics[width=.95\textwidth,trim=250 95 200 65,clip]{fig/frames/ukf-13} \\
    UKF step 130
    \captionsetup{justification=centering}
    \caption{No noise \\ in triangulation}
  \end{subfigure}%
  \hfill
  \begin{subfigure}[b]{.3\textwidth}
    \centering
%
%
\definecolor{mycolor1}{rgb}{0.48630,0.67450,0.33730}%
\definecolor{mycolor2}{rgb}{0.17250,0.43140,0.72940}%
\begin{tikzpicture}

\begin{axis}[%
xmin=-0.25,
xmax=4.25,
xlabel style={font=\color{white!15!black}},
xlabel={Variance in feature point initialization},
ymin=-1,
ymax=6,
ylabel style={font=\color{white!15!black}},
ylabel={RMSE in meters (path)},
axis background/.style={fill=white},
xmajorgrids,
ymajorgrids,
legend style={legend cell align=left, align=left, draw=white!15!black},
width=\figurewidth,
height=\figureheight
]
\addplot [color=mycolor1, line width=1.0pt]
  table[row sep=crcr]{%
0.0001	0.186245389324503\\
0.0025	0.186423808028689\\
0.01	0.186495382474899\\
0.04	0.186911092692774\\
0.16	0.230513204448009\\
0.36	0.297949559919501\\
1	0.606344663816847\\
1.69	0.990341056288841\\
2.56	1.5292555467156\\
4	2.48892287751312\\
};
\addlegendentry{EKF (linearization)}

\addplot [color=mycolor2, line width=1.0pt]
  table[row sep=crcr]{%
0.0001	0.238001848769537\\
0.0025	0.241889659995765\\
0.01	0.246974720190227\\
0.04	0.257322213174685\\
0.16	0.299306892577807\\
0.36	0.364757962714627\\
1	0.525186038278864\\
1.69	0.604482010038314\\
2.56	0.731519188380432\\
4	0.844892370788786\\
};
\addlegendentry{UKF (moment matching)}

\addplot [color=mycolor1, forget plot]
 plot [error bars/.cd, y dir=both, y explicit, error bar style={line width=0.5pt}, error mark options={line width=0.5pt, mark size=6.0pt, rotate=90}]
 table[row sep=crcr, y error plus index=2, y error minus index=3]{%
0.0001	0.186245389324503	0.0286797368191563	0.0286797368191563\\
0.0025	0.186423808028689	0.0289319142272432	0.0289319142272432\\
0.01	0.186495382474899	0.0293811792302018	0.0293811792302018\\
0.04	0.186911092692774	0.0324171247857839	0.0324171247857839\\
0.16	0.230513204448009	0.124163403793128	0.124163403793128\\
0.36	0.297949559919501	0.209043304615695	0.209043304615695\\
1	0.606344663816847	0.522681993916514	0.522681993916514\\
1.69	0.990341056288841	0.792084152578495	0.792084152578495\\
2.56	1.5292555467156	1.7201011158426	1.7201011158426\\
4	2.48892287751312	2.68447559112116	2.68447559112116\\
};
\addplot [color=mycolor2, line width=1.0pt, forget plot]
 plot [error bars/.cd, y dir=both, y explicit, error bar style={line width=1.0pt}, error mark options={line width=1.0pt, mark size=3.0pt, rotate=90}]
 table[row sep=crcr, y error plus index=2, y error minus index=3]{%
0.0001	0.238001848769537	0.0331187969735751	0.0331187969735751\\
0.0025	0.241889659995765	0.0356381830617713	0.0356381830617713\\
0.01	0.246974720190227	0.0397753326185064	0.0397753326185064\\
0.04	0.257322213174685	0.0550200391156971	0.0550200391156971\\
0.16	0.299306892577807	0.0713668469188254	0.0713668469188254\\
0.36	0.364757962714627	0.075301992914212	0.075301992914212\\
1	0.525186038278864	0.203912623989173	0.203912623989173\\
1.69	0.604482010038314	0.156916147117727	0.156916147117727\\
2.56	0.731519188380432	0.242548280174758	0.242548280174758\\
4	0.844892370788786	0.272343683073214	0.272343683073214\\
};
\end{axis}
\end{tikzpicture}%
    \caption{Normalized RMSE in path \\ ~}
  \end{subfigure}
  \hfill
  \begin{subfigure}[b]{.3\textwidth}
    \centering
%
%
\definecolor{mycolor1}{rgb}{0.48630,0.67450,0.33730}%
\definecolor{mycolor2}{rgb}{0.17250,0.43140,0.72940}%
\begin{tikzpicture}

\begin{axis}[%
xmin=-0.25,
xmax=4.25,
xlabel style={font=\color{white!15!black}},
xlabel={Variance in feature point initialization},
ymin=-0.5,
ymax=3,
ylabel style={font=\color{white!15!black}},
ylabel={RMSE in meters (map)},
axis background/.style={fill=white},
xmajorgrids,
ymajorgrids,
legend style={legend cell align=left, align=left, draw=white!15!black},
width=\figurewidth,
height=\figureheight
]
\addplot [color=mycolor1, line width=1.0pt]
  table[row sep=crcr]{%
0.0001	0.0382107680936708\\
0.0025	0.0383059390645719\\
0.01	0.038635849869834\\
0.04	0.0415602350739689\\
0.16	0.084628281168901\\
0.36	0.138986886279656\\
1	0.361111572636892\\
1.69	0.829962753890669\\
2.56	1.08408694062311\\
4	1.54251657840288\\
};
\addlegendentry{EKF (linearization)}

\addplot [color=mycolor2, line width=1.0pt]
  table[row sep=crcr]{%
0.0001	0.145217612594171\\
0.0025	0.149597390367365\\
0.01	0.168261768411524\\
0.04	0.227474939052705\\
0.16	0.316902133231553\\
0.36	0.467676506003018\\
1	0.822674837790259\\
1.69	0.968160618649451\\
2.56	1.16898605790995\\
4	1.31007596384252\\
};
\addlegendentry{UKF (moment matching)}

\addplot [color=mycolor1, forget plot]
 plot [error bars/.cd, y dir=both, y explicit, error bar style={line width=0.5pt}, error mark options={line width=0.5pt, mark size=6.0pt, rotate=90}]
 table[row sep=crcr, y error plus index=2, y error minus index=3]{%
0.0001	0.0382107680936708	0.0122461961552055	0.0122461961552055\\
0.0025	0.0383059390645719	0.0122048434182646	0.0122048434182646\\
0.01	0.038635849869834	0.0121568676762654	0.0121568676762654\\
0.04	0.0415602350739689	0.0128475649437238	0.0128475649437238\\
0.16	0.084628281168901	0.0960546613715161	0.0960546613715161\\
0.36	0.138986886279656	0.140067970244426	0.140067970244426\\
1	0.361111572636892	0.349909675835537	0.349909675835537\\
1.69	0.829962753890669	0.897482575593751	0.897482575593751\\
2.56	1.08408694062311	1.00613362188396	1.00613362188396\\
4	1.54251657840288	1.0400158135228	1.0400158135228\\
};
\addplot [color=mycolor2, line width=1.0pt, forget plot]
 plot [error bars/.cd, y dir=both, y explicit, error bar style={line width=1.0pt}, error mark options={line width=1.0pt, mark size=3.0pt, rotate=90}]
 table[row sep=crcr, y error plus index=2, y error minus index=3]{%
0.0001	0.145217612594171	0.0486871601662074	0.0486871601662074\\
0.0025	0.149597390367365	0.0436972704453663	0.0436972704453663\\
0.01	0.168261768411524	0.0507090622038128	0.0507090622038128\\
0.04	0.227474939052705	0.0740987802276704	0.0740987802276704\\
0.16	0.316902133231553	0.164095496161896	0.164095496161896\\
0.36	0.467676506003018	0.199688694450431	0.199688694450431\\
1	0.822674837790259	0.333184172930107	0.333184172930107\\
1.69	0.968160618649451	0.349733892546218	0.349733892546218\\
2.56	1.16898605790995	0.467822365855011	0.467822365855011\\
4	1.31007596384252	0.460133736602999	0.460133736602999\\
};
\end{axis}
\end{tikzpicture}%
    \caption{Normalized RMSE of map after alignment \\ ~}    
  \end{subfigure}  
  \hfill
  \begin{subfigure}[b]{.15\textwidth}
    \centering
    \includegraphics[width=.95\textwidth,trim=250 95 200 65,clip]{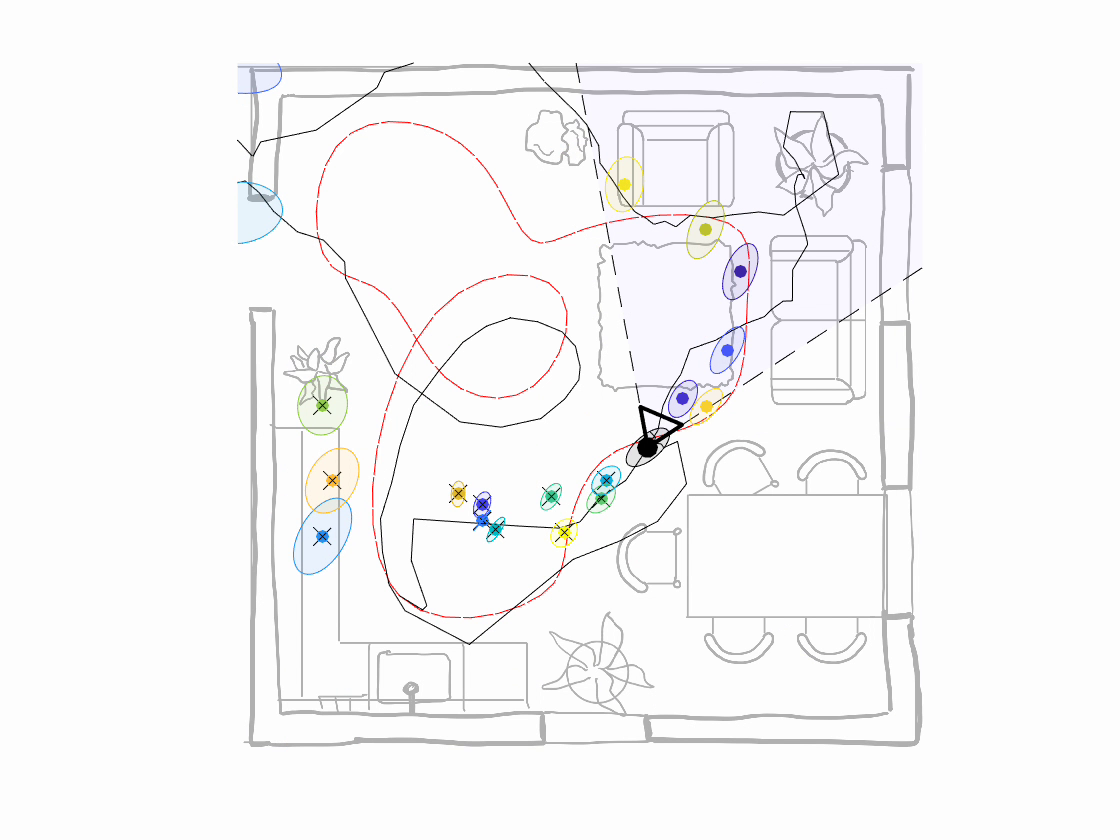} \\
    EKF step 130 \\[1em]
    \includegraphics[width=.95\textwidth,trim=250 95 200 65,clip]{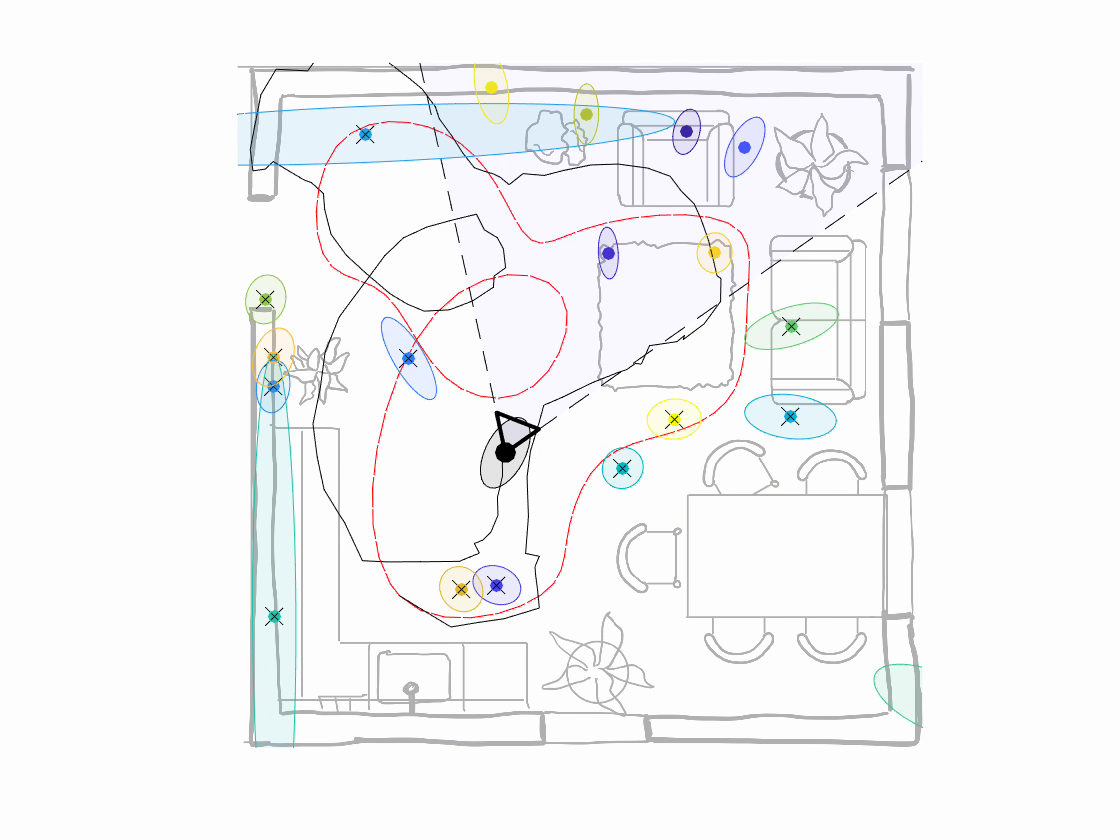} \\
    UKF step 130
    \captionsetup{justification=centering}    
    \caption{High noise \\ in triangulation}    
  \end{subfigure}%
  \vspace*{-6pt}
  \caption{Simulated SLAM under uncertainty in initial feature point triangulation. Panel (a) shows solutions for linearization (EKF) and moment matching (UKF) based SLAM under low noise. Panels (b)--(c) show the effect on RMSE over the path (b) and map feature point locations (c) after a procrustes alignment to align the scale based on the learned map (thus (c) is not very informative). In (b), we visualize the RMSE as a function of the variance of the confounding noise on the initial guess for the feature point locations. For each noise level, we visualize the mean and standard deviation error bars over 20 runs. Moment matching appears  more stable than linearization. Panel (d) shows example results under high noise levels.}
  \label{fig:noisy-slam}
  \vspace*{-6pt}
\end{figure*}

\textbf{Errors in feature point assignment:} In the first of our two perturbation studies, we corrupt the feature tracking results by randomly swapping feature point observations between nearby feature points. In \cref{fig:wrong-slam}, we show RMSE results as a function of proportion of corrupted feature point assignments. Each point corresponds to a full run of the SLAM and we visualize the moving average (solid line) and moving standard deviation (coloured patch). The model looses absolute orientation and scale in the observation model described in \cref{eq:slam-meas}, which means that the path and map can be re-scaled. Thus we perform a procrustes alignment between the ground truth map points and the estimated map and align the map and path before calculating the RMSE. The results in \cref{fig:wrong-slam} highlight directly the improved stability of the moment matching (UKF) over linearization (EKF).

\textbf{Robustness in initial triangulation:} Another recurring issue in this type of visual SLAM problems is initialization of feature point locations. In practice, the initial guess is typically vague and could be, \eg, based on triangulation from two views (small errors in pose translation/orientation lead to large error in triangulation). Because our simulation setup is simulated, we can initialize the points by taking their ground truth locations and corrupting them with Gaussian noise. In the second perturbation study we control the variance of the corrupting noise that affect the initial (prior) state mean $\vm_0$ corresponding to $\{\vp^{(i)}\}_{i=1}^{20}$.

\cref{fig:noisy-slam} shows results for controlling the feature point initialization. The evaluation setup is the same as in \cref{fig:wrong-slam}, and for each noise level, we visualize the mean and standard deviation error bars over 20 runs. The effect of deviating from the vicinity of the `good' linearization point is apparent here, and the EKF performance deteriorates quickly as the noise scale grows. Note that regardless of the scale of the corrupting noise the prior state covariance is initialized to $4^2\,\MI$, which means that the scale of the corrupting noise can be considered `small' even for large noise scales.

\subsection{Visual-inertial odometry}
\label{sec:VIO}
As a real-world experiment, we consider a visual-inertial odometry setup, where the data sources are an IMU that provides high-frequency (typically $>$100~Hz) gyroscope and accelerometer observations and a camera rigidly attached to the IMU that feeds the system with more sparsely (typically 10--60~fps) observed RGB camera images. We focus on an open-source state-of-the-art VIO method HybVIO~\cite{HybVIO}.

The VIO system state holds the translation $\vp \in \R^3$, velocity $\vv \in \R^3$, orientation quaternion $\vq \in \mathrm{SO}(3)$ (represented as $\R^4$), additive/multiplicative biases of the IMU ($\MT_k^\mathrm{a}, \vect{b}^\mathrm{a}_k, \vect{b}^\omega_k$, see below) and a trail of past camera poses (translation and orientation) $\vectb{\pi}_k^{(j)} = (\vp_{\#j}, \vq_{\#j})$ at previously observed camera frames, which are used during the measurement update for triangulating the visual features. The high-dimensional state takes the following form:
\begin{equation}
  \vx_k = (\vp_k, \vv_k, \vq_k, \MT_k^\mathrm{a}, \vect{b}^\mathrm{a}_k, \vect{b}^\omega_k, \vectb{\pi}_k^{(1)}, \ldots, \vectb{\pi}_k^{(p)}).
\end{equation}

\begin{table*}[]
\center
\caption{EuRoC MAV benchmark (RMS ATE metric with $\mathrm{SE}(3)$ alignment, in meters) in an online stereo setting. Runs where the methods diverged are denoted `---' and the statistics are computed without failed cases included.}
  \label{tbl:euroc}
  \setlength{\tabcolsep}{5.5pt} 
\begin{tabular}{lccccccccccc|c|c|c}
\toprule
\sc Method                            & MH01           & MH02           & MH03           & MH04           & MH05           & V101           & V102           & V103           & V201           & V202           & V203  & Mean          & Median        & Std  \\ \midrule
\multicolumn{1}{l|}{OKVIS \cite{leutenegger2015keyframe}}        & 0.23           & 0.15           & 0.23           & 0.32           & 0.36           & \textit{0.04}  & 0.08           & 0.13           & 0.1            & 0.17           & ---   & 0.18          & 0.16          & 0.10 \\
\multicolumn{1}{l|}{VINS-Fusion \cite{paul2017comparative}}  & 0.24           & 0.18           & 0.23           & 0.39           & 0.19           & 0.1            & 0.1            & 0.11           & 0.12           & 0.1            & ---   & 0.18          & 0.15          & 0.09 \\
\multicolumn{1}{l|}{BASALT \cite{usenko2019visual}}       & \textbf{0.07}  & \textbf{0.06}  & 0.07           & 0.13           & \textbf{0.11}  & \textit{0.04}  & 0.05           & 0.1            & \textbf{0.04}  & 0.05  & ---   & 0.07 & \textit{0.07} & 0.03 \\ 
\multicolumn{1}{l|}{HybVIO \cite{HybVIO}} & 0.088 & 0.08 & \textit{0.038} & \textbf{0.071} & \textbf{0.11} & 0.044 & \textit{0.035} & \textit{0.04} & 0.075 & \textit{0.041} & \textit{0.052} & \textit{0.061} & \textbf{0.05} & 0.02 \\
\midrule
\multicolumn{1}{l|}{HybVIO (EKF)} & 0.097          & 0.098          & 0.062 & 0.113 & 0.159          & 0.060          & 0.048 & 0.075 & 0.089          & 0.087          & ---   & 0.09          & 0.09          & 0.03 \\
\multicolumn{1}{l|}{HybVIO (UKF)} & \textit{0.078} & \textit{0.076} & \textbf{0.033} & \textit{0.074} & \textit{0.118} & \textbf{0.039} & \textbf{0.024} & \textbf{0.039} & \textit{0.05} & \textbf{0.036} & \textbf{0.046} & \textbf{0.057} & \textbf{0.05} & 0.03 \\
\bottomrule
\end{tabular}
\end{table*}

We follow~\cite{PIVO} and \cite{HybVIO}, where the IMU propagation is performed on each synchronized pair $(\vomega_k, \va_k)$ of gyroscope and accelerometer samples as an EKF prediction step of the form in \cref{eq:dynamics} with $\vepsilon_k \sim \mathrm{N}(\vzero, \MQ \Delta t_k)$. The function $\vf_k(\cdot,\cdot)$ updates the pose and velocity by the mechanization equation
\begin{equation}
  \label{eq:imu-propagation}
  \begin{pmatrix}
    \vect{p}_k \\ \vect{v}_k \\ \vect{q}_k
  \end{pmatrix}
  =
  \begin{pmatrix} %
    \vect{p}_{k-1} + \vect{v}_{k-1}\Delta t_k \\
    \vect{v}_{k-1} + [\vect{q}_k (\tilde{\vect{a}}_k + \vectb{\varepsilon}^\mathrm{a}_k) \vect{q}_k^\star - \vect{g}] \Delta t_k \\
    \vectb{\Omega}[(\tilde{\vectb{\omega}}_k + \vectb{\varepsilon}^\omega_k) \Delta t_k] \vect{q}_{k-1}
  \end{pmatrix},
\end{equation}
where the bias-corrected IMU measurements are computed as $\tilde{\vect{a}}_k = \MT_k^\mathrm{a} \, \va_k - \vb^\mathrm{a}_k$ and $\tilde{\vomega}_k = \vomega_k - \vb^\omega_k$. In the model, the multiplicative correction $\vect{T}_k^\mathrm{a} \in \R^{3\times3}$ is a diagonal matrix. Note that, contrary to the approach used in~\cite{MSCKF}, the formulation above does not involve linearization errors that could cause the orientation quaternion to lose its unit length, since $\vectb{\Omega}[\cdot] \in \R^{4 \times 4}$ (\cf~\cite{Titterton+Weston:2004}) is an orthogonal matrix. For the visual feature tracking and measurement models we follow the approach in the HybVIO~\cite{HybVIO} paper. For simplicity, we do not implement the IMU bias mean-reversion model presented in the HybVIO paper.

We take the open-source C++ implementation of HybVIO (\url{https://github.com/SpectacularAI/HybVIO}) as our starting point and modify the time update (prediction) step following the presentation in \cref{sec:methods}. For direct comparison between the EKF and the UKF, we do not modify the visual update step, but still use an EKF update in the experiments. The reasons for this is the additional outlier checks and innovation tests that have been implemented for the EKF. Accounting for these under the UKF update would require extensive parameter tuning, which would make direct and fair comparison between methods difficult. By modifying the prediction step only, no {\it ad~hoc} tuning is required for comparing the methods. We still speculate, that an UKF-style update could carry over similar benefits as shown in the experiments in the previous section, though with additional computational load.

We evaluate the methods on real data from the widely used EuRoC MAV benchmark \cite{EuRoC}. This benchmark contains 11 sets of drone flying data indoors, where the drone has a stereo camera pair and an IMU rigidly attached to its frame. The data sets are captured in two different indoor spaces, and they are of increasing difficulty (this also shows in the results table). The benchmark is typically considered separately for mono vs.\ stereo and online vs.\ batch ($\sim$filtering/smoothing) processing. We test the modified VIO method in the online stereo setting, which corresponds to real-time estimation of the drone pose using the data seen so far.

In \cref{tbl:euroc}, we report results on the EuRoC MAV benchmark (RMS ATE metric with $\mathrm{SE}(3)$ alignment, in meters) in the online stereo setting. Runs where the methods diverged are denoted by `---' (for the best methods, this typically only happens in the challenging V203 data set). The statistics on the right are calculated without failed cases included, which gives them an upper hand. We compare to other recent state-of-the-art methods for the online benchmark (for other methods and categories, see \cite{HybVIO}).

Because we did not implement the mean-reverting random walk model for the IMU biases and we also change the initial state covariances for the velocity and orientation ($10^{-6}$ and $10^{-3}$, respectively), we run HybVIO both in EKF and UKF mode for comparison. These results are reported on the bottom rows. The EKF run diverged in this case for the challenging V203 set. Overall, HybVIO with UKF (moment matching) predictions shows excellent results, also beating previous HybVIO results.

\section{Discussion and Conclusion}
\noindent
In this paper, we revisited the assumed density view to non-linear state estimation and provided an alternative approach to established VIO and visual SLAM through this viewpoint. In parts, our approach here was empirical with focus on showing practical results both in a simulated visual SLAM setting and on real-world drone flying benchmark data.

In both experiment setups the results are almost surprisingly consistent: the moment matching (UKF) seems to improve robustness and results in all experiments out-of-the-box. Given known properties and discussion in the extensive literature related to non-linear filtering, the results are hardly surprising, but rather support the existing knowledge of improved robustness over EKF. The differences should become more pronounced once deviating more from the vicinity of the linearization point, which also appears clear in the results we provide. \looseness-1

In the visual SLAM experiment in \cref{sec:EKF-SLAM}, we did not consider any additional outlier rejection methods to stabilize the performance. In practice, these would be implemented through, \eg, innovation tests. In \cref{sec:VIO}, we did minimal parameter tuning on top of the HybVIO implementation which has been parameter-wise optimized for the EKF predictions and updates. We speculate that many of the parameters that have a physical interpretation (such as noise scales for velocities, \etc) could be given more sensible values, when moment matching is used instead of direct linearization. Yet, we have to underline that switching away from EKF typically comes at a cost. In the experiments here we require multiple evaluations of the non-linear dynamics and observation models, which can be considered costly at least on embedded hardware, even if the practical runtime for the methods evaluated on a MacBook pro laptop did slow down by a factor of $\sim$two (still ensuring real-time performance). On the other hand, many of the operations can be parallelized. Also hybrid methods, where one would start off with a UKF-type of model, and switch to an EKF when the estimation has converges to track well.

Given the promising results from moment matching, which relates to a single-sweep of expectation propagation (EP) in machine learning, we would expect other related methods to further improve the inference. For example, Chang \etal~\cite{chang2020fast} studied sequential variants of conjugate-computation variational inference that could further improve inference aspects in these models. Yet, the latent state variables in the two model families in the experiments have strong inner couplings. Couplings of this kind have been discussed, \eg, in \cite{wilkinson2021bayes}, which could improve upon the performance in these cases.

We see promise in considering alternative non-linear filtering methods to EKF for inference in this type of models. The results signal that better-tailored inference methods can directly improve performance in visual SLAM and VIO. This is directly highlighted by the real-world benchmark results as well: In the online stereo category for the EuRoC benchmark, we report the best real-time accuracy ever published for this widely-used test data.

\section*{Acknowledgements}
\noindent
We acknowledge the computational resources provided by the Aalto Science-IT project and CSC -- IT Center for Science, and funding from Academy of Finland (grants 339730 and 324345) and Finnish Center for Artificial Intelligence (FCAI). The authors wish to thank Spectacular~AI for valuable technical input. \looseness-1

{\small
\bibliographystyle{ieee}

}

\end{document}